\documentclass{article}

\usepackage{microtype}
\usepackage{graphicx}
\usepackage{subfigure}
\usepackage{booktabs}
\usepackage{nicefrac}
\usepackage{hyperref}

\usepackage[accepted]{icml2025}

\usepackage{amsmath}
\usepackage{amssymb}
\usepackage{mathtools}
\usepackage{amsthm}
\usepackage{stmaryrd}
\usepackage[capitalize,noabbrev]{cleveref}

\theoremstyle{plain}
\newtheorem{theorem}{Theorem}[section]
\newtheorem{proposition}[theorem]{Proposition}

\theoremstyle{definition}
\newtheorem{definition}[theorem]{Definition}

\theoremstyle{remark}

\usepackage[textsize=tiny]{todonotes}

\icmltitlerunning{Antithetic Sampling for Top-k Shapley Identification}

\begin{document}

\twocolumn[
\icmltitle{Antithetic Sampling for Top-k Shapley Identification}

\icmlsetsymbol{equal}{*}

\begin{icmlauthorlist}
\icmlauthor{Patrick Kolpaczki}{equal,lmu,mcml}
\icmlauthor{Tim Nielen}{equal,lmu}
\icmlauthor{Eyke H\"ullermeier}{lmu,mcml}
\end{icmlauthorlist}

\icmlaffiliation{lmu}{LMU Munich}
\icmlaffiliation{mcml}{Munich Center for Machine Learning}

\icmlcorrespondingauthor{Patrick Kolpaczki}{patrick.kolpaczki@ifi.lmu.de}

\icmlkeywords{Shapley Value, Game Theory, Explainable AI}

\vskip 0.3in
]

\printAffiliationsAndNotice{\icmlEqualContribution}

\begin{abstract}
    Additive feature explanations rely primarily on game-theoretic notions such as the Shapley value by viewing features as cooperating players.
The Shapley value's popularity in and outside of explainable AI stems from its axiomatic uniqueness.
However, its computational complexity severely limits practicability.
Most works investigate the uniform approximation of all features' Shapley values, needlessly consuming samples for insignificant features.
In contrast, identifying the $k$ most important features can already be sufficiently insightful and yields the potential to leverage algorithmic opportunities connected to the field of multi-armed bandits.
We propose \emph{Comparable Marginal Contributions Sampling} (CMCS), a method for the \emph{top-$k$ identification problem} utilizing a new sampling scheme taking advantage of correlated observations.
We conduct experiments to showcase the efficacy of our method in compared to competitive baselines.
Our empirical findings reveal that estimation quality for the \emph{approximate-all problem} does not necessarily transfer to \emph{top-$k$ identification} and vice~versa.
\end{abstract}

\section{Introduction}
\label{sec:introduction}

The fast-paced development of artificial intelligence
poses a double-edged sword.
Obviously on one hand, machine learning models have significantly improved in prediction performance, most famously demonstrated by deep learning models.
But, on the other hand, their required complexity to exhibit these capabilities comes at a price.
Human users face concerning challenges comprehending the decision-making of such models that appear to be increasingly opaque.
The field of explainable AI \cite{Vilone.2021,Molnar.2022} offers a simple yet popular approach to regain understanding and shed light onto these black box models by means of \emph{additive feature explanations} \cite{Doumard.2022}.
Probing a model's behavior to input, this explanation method assigns importance scores to the utilized features.
Depending on the explanandum of interest, each score can be interpreted as the feature's impact on the models' prediction for a particular instance or its generalization performance.

The Shapley value \cite{Shapley.1953} has emerged as a prominent mechanism to assign scores.
Taking a game-theoretic perspective, each feature is viewed as a player in a \emph{cooperative game} in which the players can form coalitions and reap a \emph{collective benefit} by solving a task together.
For instance, a coalition representing a feature subset can be rewarded with the generalization performance of the to be explained model using only that subset.
Posing the omnipresent question of how to divide in equitable manner the collective benefit that all players jointly achieve, reduces the search for feature importance scores to a \emph{fair-division problem}.
The Shapley value is the unique solution to fulfill certain desiderata which arguably capture an intuitive notion of fairness \cite{Shapley.1953}.
The marginal contributions of a player to all coalitions, denoting the increase in collective benefit when joining a coalition, are taken into a weighted sum by the Shapley value.

It has been extensively applied for \emph{local explanations}, dividing the~\mbox{prediction} value \cite{Lundberg.2017}, and \emph{global explanations} that divide prediction performance \cite{Covert.2020}.
In addition to providing understanding, other works proposed to utilize it for the selection of machine learning entities such as features \cite{Cohen.2007,Wang.2024},
datapoints \cite{Ghorbani.2019},
neurons in deep neural networks \cite{Ghorbani.2020}, or base learners in ensembles \cite{Rozemberczki.2021}.
We refer to \cite{Rozemberczki.2022} for an overview of its applications in machine learning.
Unfortunately, the complexity of the Shapley value poses a serious limitation: its calculation encompasses all coalitions within the exponentially growing power set of players.
Hence, the exact computation of the Shapley value is quickly doomed for even moderate feature numbers.
Ergo, the research branch of estimating the \mbox{Shapley} value has sparked notable interest, in particular the challenge of precisely approximating the Shapley values of all players known as the \emph{approximate-all~problem}. 

However, often the exact importance scores just serve as a means to find the most influential features, be it for explanation or preselection \cite{Cohen.2007,Wang.2024}, and are not particularly relevant themselves.
Hence, we advocate for the \emph{top-$k$ identification problem} \cite{Kolpaczki.2021} in which an approximation algorithm's goal is to identify the $k$ players with highest Shapley values, without having to return precise estimates.
This incentivizes to forego and sacrifice precision of players' estimates for whom reliable predictions of top-$k$ membership already manifest during runtime.
Instead, the available samples, reflecting finite computational power at disposal, are better spent on players on the verge of belonging to the top-$k$ in order to speed up the segregation of top-$k$ players from~the~rest. 

\paragraph{Contribution.}
We propose with \emph{Comparable Marginal Contributions Sampling} (CMCS), Greedy CMCS, and CMCS$@$K novel top-$k$ identification algorithms for the Shapley value.
More specifically, our contributions are: 

\begin{itemize}
    \item We present a new representation of the Shapley value based on an altered notion of marginal contribution and leverage it to develop CMCS.
    On the theoretical basis of antithetic sampling, we underpin the intuition behind utilizing correlated observations especially for top-$k$ identification.
    \item 
    Moreover, with Greedy CMCS and CMCS$@$K we propose multi-armed bandit-inspired enhancements.
    Our proposed algorithms are model-agnostic and applicable to any cooperative game independent of the domain of interest.
    \item Lastly, we observe how empirical performance does not directly translate~from the approximate-all to the top-$k$ identification problem.
    Depending~on~the task, different algorithms are favorable and a conscious choice~is~advisable.
\end{itemize}

\section{Related Work}
\label{sec:related_work}

The problem of precisely approximating all players' Shapley values has been extensively investigated.
Since the Shapley value is a weighted average of a player's marginal contributions, methods that conduct mean estimation form a popular class of approximation algorithms.
Most of these sample marginal contributions as performed by \emph{ApproShapley} \cite{Castro.2009}.
Many variance reduction techniques, that increase the estimates' convergence speed, have been incorporated: stratification \cite{Maleki.2013,OBrien.2015,Castro.2017,vanCampen.2018,Okhrati.2020,Burgess.2021}, antithetic sampling \cite{Illes.2019,Mitchell.2022}, and control variates \cite{Goldwasser.2024}.
\mbox{Departing} from the notion of marginal contributions, other methods view the Shapley value as a composition of coalition values and sample these instead for mean estimation \cite{Covert.2019a,Kolpaczki.2024a,Kolpaczki.2024b}.
A different class of methods does not approximate Shapley values directly, but fits a parametrized surrogate game via sampling.
As the surrogate game represents the game of interest increasingly more faithful, its own Shapley values become better estimates.
Due to the surrogate game's highly restrictive structure these
can be obtained in polynomial time.
\emph{KernelSHAP} \cite{Lundberg.2017} is the most prominent member of this class with succeeding extensions \cite{Covert.2021,Pelegrina.2025}.
See \cite{Chen.2023} for an overview of further methods for feature attribution and specific model~classes.

First to consider the top-$k$ identification problem for Shapley values were \citet{Ramasuri.2008} by simply returning the players with the highest estimates effectively computed by \emph{ApproShapley} \cite{Castro.2009}.
This straightforward reduction of top-$k$ identification to the approximate-all problem can be realized with any approximation algorithm.
\citet{Kolpaczki.2021} establish a connection to the field of multi-armed bandits \cite{Lattimore.2020} and thus open the door to further algorithmic opportunities that top-$k$ identification has to offer.
Here, pulling an arm of a slot machine metaphorically captures the draw of a sample from a distribution.
Usually, one is interested in maximizing the cumulative random reward obtained from sequentially playing the multi-armed slot machine or finding the arm with highest mean reward.
Modeling each player as an arm and its reward distribution to be the player's marginal contributions distributed according to their weights within the Shapley value \cite{Kolpaczki.2021}, facilitates the usage of bandit algorithms to find the $k$ distributions with highest mean values which represent the players' Shapely values.
The inherent trade-off between constantly collecting information from all arms to avoid falling victim to the estimates' stochasticity and selecting only those players that promise the most information gain to correctly predict top-$k$ membership, constitutes the well-known exploration-exploitation dilemma.

Bandit algorithms such as \emph{Gap-E} \cite{Gabillon.2011} and \emph{Border Uncertainty Sampling (BUS)} \cite{Kolpaczki.2021} tackle it by greedily selecting the next arm to pull as the one that maximizes a selection criterion which combines the uncertainty of top-$k$ membership and its sample number.
In contrast \emph{Successive Accepts and Rejects (SAR)} \cite{Bubeck.2013} phase-wise eliminates arms whose top-$k$ membership can be reliably predicted.
\emph{SHAP$@$K} \cite{Kariyappa.2024} employs an alternative greedy selection criterion based on confidence intervals for the players' estimates.
In each round, samples are taken from two players, one from the currently predicted top-$k$ and one outside of them, with the highest overlap in confidence intervals.
The overlap is interpreted~as~the likelihood that the pair is mistakenly partitioned and should~be~swapped~instead.
\section{The Top-$k$ Identification Problem}
\label{sec:top_k}

We introduce cooperative games and the Shapley value formally in \cref{subsec:games_and_SV}, and briefly after present the widely studied problem of approximating all players' Shapley values in a cooperative game \cref{subsec:approximate_all}.
On that basis, we introduce the problem of identifying the top-$k$ players with the highest Shapley values in \cref{subsec:top_k_identification} and distinguish it from the former by highlighting decisive differences in performance measures which will prepare our theoretical findings and arising methodological avenues alluded to in~\cref{sec:topk_vs_mse}.

\subsection{Cooperative Games and the Shapley Value}
\label{subsec:games_and_SV}

A cooperative game $(\mathcal{N}, \nu)$ consists of a \emph{player set} $\mathcal{N} = \{1,\ldots,n\}$ and a \emph{value function} $\nu : \mathcal{P}(\mathcal{N}) \to \mathbb{R}$ that maps each subset $S \subseteq \mathcal{N}$ to a real-valued worth.
The players in $\mathcal{N}$ can cooperate by forming \emph{coalitions} in order to achieve a goal.
A coalition is represented by a subset $S$ of $\mathcal{N}$ that includes exactly all players which join the coalition.
The formation of a coalition resolves in the (partial) fulfillment of the goal and a collective benefit $\nu(S)$ disbursed to the coalition which we call the \emph{worth} of that coalition.
The empty set has no worth, i.e.\ $\nu(\emptyset)=0$.
The abstractness of this notion offers a certain versatility in modeling many cooperative scenarios.
In the context of feature explanations for example, each player represents a feature and the formation of a coalition is interpreted to express that a model or learner uses only that feature subset and discards those features absent in the coalition.
Depending on the desired explanation type, the prediction value for a datapoint of interest or an observed behavior of the model over multiple instances, for example generalization performance on a test set, is commonly taken as the worth of a feature subset.

A central problem revolving around cooperative games is the question of how to split the collective benefit that all players achieve together among them.
More precisely, which share $\phi_i$ of the \emph{grand coalition's} worth $\nu(\mathcal{N})$ should each player $i \in \mathcal{N}$ receive?
A common demand is that these payouts $\phi$ are to be fair and reflect the contribution that each player provides to the fulfillment of the goal.
Guided by this rationale, the Shapley value \cite{Shapley.1953} offers a popular solution by assigning each player $i$ the payoff
\begin{equation} \label{eq:Shapley}
    \phi_i = \sum\limits_{S \subseteq \mathcal{N} \setminus \{i\}} \frac{1}{n \binom{n-1}{|S|}} \cdot [\nu(S \cup \{i\}) - \nu(S)] \, .
\end{equation}
The difference in worth $\Delta_i(S) := \nu(S \cup \{i\}) -\nu(S)$ is known as \emph{marginal contribution} and reflects the increase in collective benefit that $i$ causes by joining the coalition $S$.
The reason for the Shapley value's popularity lies within its axiomatic justification.
It is the unique payoff distribution to simultaneously satisfy the four axioms, symmetry, linearity, efficiency, and dummy player \cite{Shapley.1953}, which capture an intuitive notion of fairness in light of the faced fair division problem.
Despite this appeal, the Shapley value comes with a severe drawback.
The number of coalition values contained in its summation grows exponentially w.r.t.\ the number of players $n$ in the game.
In fact, its exact calculation is provably NP-hard \cite{Deng.1994} if no further assumption on the structure of $\nu$ is made, and as a consequence, the Shapley value becomes practically intractable for datasets with even medium-sized feature numbers.
This issue necessitates the precise estimation of Shapley values to provide accurate explanations.

\subsection{Approximating all Shapley Values}
\label{subsec:approximate_all}

Within the \emph{approximate-all problem}, the objective of an approximation algorithm $\mathcal{A}$ is to precisely estimate the Shapley values $\phi = (\phi_1,\ldots,\phi_n)$ of all players by means of estimates $\hat\phi = (\hat\phi_1,\ldots,\hat\phi_n)$ for a given cooperative game $(\mathcal{N}, \nu)$.
We consider the \emph{fixed-budget} setting in which the number of times $\mathcal{A}$ can access $\nu$ to evaluate the worth $\nu(S)$ of a coalition $S$ of its choice is limited by a budget $T \in \mathbb{N}$.
Thus, $\mathcal{A}$ can sequentially retrieve the worth of $T$ many, possibly duplicate, coalitions to construct its estimate $\hat\phi$.
This captures the limitation in time, computational resources, or monetary units that a practical user is facing to avoid falling victim to the exact computation's complexity.
Furthermore, it is motivated by the observation that the access to $\nu$ poses a common bottleneck, by performing inference of complex models or re-training on large data, instead of the negligible arithmetic operations of $\mathcal{A}$.

Since $\mathcal{A}$ potentially uses randomization, for instance by drawing samples and evaluating random coalitions, the comparison of $\hat\phi$ and $\phi$ needs to incorporate this randomness to judge the approximation quality.
In light of this, the expected \emph{mean squared error} is a wide-spread measure of approximation quality that is to be minimized by $\mathcal{A}$:
\begin{equation} \label{eq:MSE}
    \mathbb{E}[\text{MSE}] := \frac{1}{n} \sum_{i \in \mathcal{N}} \mathbb{E} \left[ \left( \phi_i-\hat\phi_i \right)^2 \right] \, .
\end{equation}

\subsection{Identifying Top-$k$ Players: A Subtle but Significant Difference}
\label{subsec:top_k_identification}

Instead of estimating the exact Shapley values of \emph{all} players, of which many might be similar and insignificant, one could be interested in just finding the players that possess the highest Shapley values, with the particular values being incidental.
More precisely, in the \emph{top-$k$ identification problem} (TkIP) an approximation algorithm $\mathcal{A}$ is confronted with the task of returning an estimate $\hat{\mathcal{K}} \subseteq \mathcal{N}$ of the coalition $\mathcal{K}^*$ with given size $k \in [n] := \{1,\ldots,n\}$ that contains the players with the highest Shapley values in the game $(\mathcal{N},\nu)$.
We consider again the fixed-budget setting with budget $T$.

However, $\mathcal{K}^*$ is not necessarily unique as players may share the same Shapley value.
We restrain from any assumptions on the value function $\nu$ and will thus present notions and measures capable of handling the ambiguity of $\mathcal{K}^*$.
We call a coalition $\mathcal{K} \subseteq \mathcal{N}$ of $k$ many players \emph{eligible} if the sum of Shapley values associated to the players in $\mathcal{K}$ is maximal:
\begin{equation} \label{eq:eligble}
    \sum\limits_{i \in \mathcal{K}} \phi_i = \max_{S \subseteq \mathcal{N} : |S| = k} \sum\limits_{i \in S} \phi_i \, .
\end{equation}
We denote by $\mathcal{E}_k \subseteq \mathcal{P}(\mathcal{N})$ the set of all eligible coalitions.
Any eligible estimate $\hat{\mathcal{K}}$ is correct and $\mathcal{A}$ should not be punished for it.
Note that for distinct Shapley values we have $\mathcal{E}_k = \{\mathcal{K}^*\}$.
In the following, we give in a first step precision measures (to be maximized) and error measures (to be minimized) for $\hat{\mathcal{K}}$ given $\mathcal{E}_k$ and extend them in a second step to the randomness of $\mathcal{A}$.
A straightforward way to judge the quality of an estimate $\mathcal{K}$ is the \emph{binary precision} \cite{Kolpaczki.2021}
\begin{equation} \label{eq:binary_precision}
    \psi_{\text{bin}}(\hat{\mathcal{K}}) :=
    \begin{cases}
        1 & \text{if } \hat{\mathcal{K}} \in \mathcal{E}_k \\
        0 & \text{otherwise}
    \end{cases}
\end{equation}
that maximally punishes every wrongly included player in $\hat{\mathcal{K}}$.
In order to further differentiate estimates that are close to being eligible from ones that have little overlap with an eligible coalition, we introduce the \emph{ratio precision}
\begin{equation} \label{eq:ratio_precision}
    \psi_{\text{rat}}(\hat{\mathcal{K}}) := \frac{1}{k} \max\limits_{\mathcal{K} \in \mathcal{E}_k} |\mathcal{K} \cap \hat{\mathcal{K}} |
\end{equation}
which measures the percentage of correctly identified players in $\hat{\mathcal{K}}$ by counting how many players can remain in $\hat{\mathcal{K}}$ after swapping with players from $\mathcal{N} \setminus \hat{\mathcal{K}}$ to form an eligible coalition.
It serves as a gradual but still discrete refinement of the binary precision with both measures assigning values in the unit interval $[0,1]$.
Let $\phi_{k^*} := \min\nolimits_{\mathcal{K} \in \mathcal{E}_k} \min\nolimits_{i \in \mathcal{K}} \phi_i$ be the minimal Shapley value in any eligible coalition.
Obviously, it is the minimal value for all coalitions in $\mathcal{E}_k$.
\citet{Kariyappa.2024} propose the \emph{inclusion-exclusion error} which is the smallest $\varepsilon > 0$ that fulfills
\begin{equation} \label{eq:inclusion_exclusion}
    \underbrace{\phi_i \geq \phi_{k^*} - \varepsilon}_{\text{inclusion}} \hspace{0.3cm} \text{and} \hspace{0.3cm} \underbrace{\phi_j \leq \phi_{k^*} + \varepsilon}_{\text{exclusion}}
\end{equation}
for all $i \in \hat{\mathcal{K}}$ and all $j \in \mathcal{N} \setminus \hat{\mathcal{K}}$:
\begin{multline} \label{eq:inclusion_error}
    \rho_{\text{inc+exc}} := \inf \{ \varepsilon \in \mathbb{R}^{\geq 0} \mid \forall i \in \hat{\mathcal{K}} : \phi_i \geq \phi_{k^*} - \varepsilon, \\ 
    \forall j \in \mathcal{N} \setminus \hat{\mathcal{K}} : \phi_j \leq \phi_{k^*} + \varepsilon \} \, .
\end{multline}
In simple terms, it measures how much the sum of Shapley values associated with $\hat{\mathcal{K}}$ can increase at least or that of $\mathcal{N} \setminus \hat{\mathcal{K}}$ can decrease by swapping a single player between them.
To account for the randomness of $\mathcal{A}$, effectively turning $\hat{\mathcal{K}}$ into a random variable, the expectation of each measure poses a reasonable option just as in \cref{subsec:approximate_all}.
Worth mentioning is that $\mathbb{E}[\psi_{\text{bin}}(\hat{\mathcal{K}})]$ turns out to be the probability that $\mathcal{A}$ flawlessly solves the top-$k$ identification problem.
\citet{Kariyappa.2024} resort to probably approximate correct (PAC) learning.
Specifically for the inclusion-exclusion error they call $\mathcal{A}$ for $\delta \in [0,1]$ an $(\epsilon, \delta)$-PAC learner if
\begin{equation} \label{eq:PAC}
    \mathbb{P}( \rho_{\text{inc+exc}}(\hat{\mathcal{K}}) \leq \varepsilon) \geq 1 - \delta
\end{equation}
holds after $\mathcal{A}$ terminates on its own with unlimited budget at disposal.
Obviously, any algorithm for the approximate-all problem can be translated to top-$k$ identification by simply returning the $k$ players with the highest estimates. 

\section{The Opportunity of Correlated Observations}
\label{sec:topk_vs_mse}

The two problems of approximating all players and top-$k$ identification differ in goal and quality measures, hence they also incentivize different sampling schemes.
It is the aim of our work to emphasize and draw attention to our observation that the role of correlated samples between players plays a fundamental role for the top-$k$ identification problem, whereas this is not the case for the approximate-all problem.
We demonstrate this at the example of a simple and special class of approximation algorithms that can solve both problem statements.
We call an algorithm $\mathcal{A}$ an \emph{unbiased equifrequent player-wise independent sampler} if it samples marginal contributions for all players in $M$ many rounds.
In each round $m \in \{1,\ldots,M\}$ $\mathcal{A}$ draws $n$ coalitions $S_1^{(m)},\ldots,S_n^{(m)}$, one for each $i \in \mathcal{N}$, according to a fixed joint probability distribution over $\mathcal{P}(\mathcal{N} \setminus \{1\}) \times \ldots \times \mathcal{P}(\mathcal{N} \setminus \{n\})$ with marginal distribution
\begin{equation} \label{eq:marginal_distribution}
    \mathbb{P} \left( S_i^{(m)} = S \right) = \frac{1}{n \cdot \binom{n-1}{|S|}}
\end{equation}
for each $i \in \mathcal{N}$.
Note that this implies $\mathbb{E}[\Delta_i(S_i^{(m)})] = \phi_i$ for all players.
Further, the samples are independent between rounds and $\mathcal{A}$ aggregates the samples of each player to an estimate of its Shapley value $\hat\phi_i$ by taking the mean of their resulting marginal contributions, i.e.
\begin{equation} \label{eq:sample_mean}
    \hat\phi_i = \frac{1}{M} \sum\limits_{m=1}^M \Delta_i \left( S_i^{(m)} \right) \, ,
\end{equation}
which is an unbiased estimate of $\phi_i$.
For the approximate-all problem $\mathcal{A}$ simply returns these estimates and for identifying the top-$k$ players it returns the set of $k$ players $\hat{\mathcal{K}}$ that yield the highest estimates $\hat\phi_i$.
Ties can be solved arbitrarily.
A well-known member of this class of approximation algorithms is \emph{ApproShapley} proposed by \citet{Castro.2009}.
For the approximate-all problem one can quickly derive the expected mean squared error of $\mathcal{A}$ to be
\begin{equation} \label{eq:MSE_result}
    \mathbb{E} [\text{MSE}] = \frac{1}{n M} \sum\limits_{i \in \mathcal{N}} \sigma_i^2 \, ,
\end{equation}
where $\sigma_i^2 := \mathbb{V}[\Delta_i(S_i^{(m)})]$ denotes the variance of player $i$'s marginal contributions.
The expected MSE decreases for a growing number of samples $M$ and the sum of variances $\sigma_i^2$ can be seen as a constant property of the game $(\mathcal{N}, \nu)$ that is independent of $\mathcal{A}$.
In contrast, turning to top-$k$ identification, we show the emergence of another quantity in \cref{the:inc_exc_bound} if one considers the inclusion-exclusion error.
Let $\mathbb{K}_{\varepsilon} := \{ \mathcal{K} \subseteq \mathcal{N} \mid |\mathcal{K}|=k, \rho_{\text{inc+exc}}(\mathcal{K}) \leq \varepsilon \}$ for any $\varepsilon \in \mathbb{R}^{\geq 0}$.
The central limit theorem can be applied within our considered class and thus we assume each $\sqrt{M} ((\hat\phi_i - \hat\phi_j) - (\phi_i - \phi_j))$ to be normally distributed.
\begin{theorem} \label{the:inc_exc_bound}
    Every unbiased equifrequent player-wise independent sampler $\mathcal{A}$ for the top-$k$ identification problem returns for any cooperative game~$(\mathcal{N},\nu)$~an estimate $\hat{\mathcal{K}}$ with inclusion-exclusion error of at most $\varepsilon \geq 0$ with probability~at~least
    \begin{equation*} \label{eq:lower-bound}
        \mathbb{P} (\hat{\mathcal{K}} \in \mathbb{K}_{\varepsilon}) \geq \sum_{\mathcal{K} \in \mathbb{K}_\epsilon} \left[ 1-\sum_{\substack{i \in \mathcal{K} \\ j \in \mathcal{N} \setminus \mathcal{K}}} \Phi \left( \sqrt{M} \frac{\phi_j -\phi_i}{\sigma_{i,j}} \right) \right] \, ,
    \end{equation*}
    where $\sigma_{i,j}^2 := \mathbb{V}[\Delta_i(S_i^{(m)}) - \Delta_j(S_j^{(m)})]$ and $\Phi$ denotes the standard normal cumulative distribution function.
\end{theorem}
The proof is given in \cref{app:lower_bound}.
Notice the difference to \cref{eq:MSE_result} for approximating all Shapley values.
The MSE directly reflects the change of each single player's estimate $\hat\phi_i$, but in contrast, for identifying top-$k$ an estimate may change arbitrarily as long as the partitioning of the players into top-$k$ and outside of top-$k$ stays the same.

For most pairs $i,j$ with $i \in \mathcal{K}$ and $j \in \mathcal{N} \setminus \mathcal{K}$ of a coalition $\mathcal{K} \in \mathbb{K}_{\varepsilon}$ with sufficiently small $\epsilon$, it holds $\phi_i > \phi_j$. 
Thus, for a fixed game $(\mathcal{N}, \nu)$ and fixed budget $T$, the lower bound in \cref{the:inc_exc_bound} should favorably increase if $\sigma_{i,j}$ decreases which can be influenced by $\mathcal{A}$ due to the allowed flexibility in its sampling scheme.
Note that $\mathcal{A}$ is only restricted in the marginal contribution of each $S_i^{(m)}$ but not in the joint distribution of $S_1^{(m)},\ldots,S_n^{(m)}$.
In fact, the variance of the difference between marginal contributions decomposes to
\begin{equation} \label{eq:variance_dif}
    \sigma_{i,j}^2 = \sigma_i^2 + \sigma_j^2 - 2 \text{Cov}\left( \Delta_i \left( S_i^{(m)} \right), \Delta_j \left( S_j^{(m)} \right) \right) \, .
\end{equation}
Consequently, an increased covariance between sampled marginal contributions of top-$k$ players and bottom players improves our lower bound.
Leveraging the impact of covariance shown by \cref{eq:variance_dif} in the sampling procedure is generally known as \emph{antithetic sampling}, a variance reduction technique for Monte Carlo methods to which our class belongs.
Our considered class of approximation algorithms does not impose any restrictions on the contained covariance between marginal contributions sampled within the same round $m$.
We interpret this as degrees of freedom to shape the sampling distribution.
Striving towards more reliable estimates $\hat{\mathcal{K}}$, we propose in \cref{sec:method} an approach based on the suspected improvement that positively correlated observations promise.

\section{Antithetic Sampling Approach}
\label{sec:method}

Motivated by \cref{sec:topk_vs_mse}, we develop in \cref{subsec:CMCS} \emph{Comparable Marginal Contributions Sampling} (CMCS), a budget-efficient antithetic sampling procedure that naturally yields correlated observations applicable for both problem statements.
We take inspiration from \cite{Kolpaczki.2021,Kariyappa.2024} and extend CMCS with a greedy selection criterion in \cref{subsec:greedy}, deciding from which players to sample from, to exploit opportunities that top-$k$ identification offers.
\begin{figure*}[h]
    \begin{center}
        \includegraphics[width=0.9\textwidth]{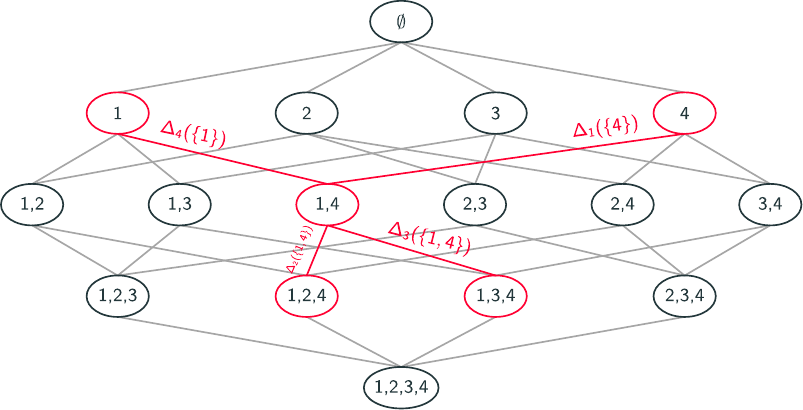}
        \caption{
            A cooperative game spans a lattice with each coalition $S \subseteq \mathcal{N}$ forming a node and each marginal contribution $\Delta_i(S)$ being represented by an edge between $S$ and $S \cup \{i\}$, exemplified here for $\mathcal{N}=\{1,2,3,4\}$.
            CMCS draws a random coalition $S$ and computes the extended marginal contributions $\Delta'_i(S) = \Delta_i(S \setminus \{i\})$ of all players $i \in \mathcal{N}$.
            For $n=4$ it evaluates five coalitions and retrieves four marginal contributions.
        }
        \label{fig:lattice}
    \end{center}
    \vskip -0.2in
\end{figure*}

\subsection{Sampling Comparable Marginal Contributions}
\label{subsec:CMCS}

We start by observing that the sampling of marginal contributions can be designed to consume less than two evaluations of $\nu$ per sample.
In fact, the budget restriction $T$ is not coupled to the evaluation of marginal contributions as atomic units but single accesses to $\nu$.
Instead of separately evaluating $\nu(S)$ and $\nu(S \cup \{i\})$ for each $\Delta_i(S)$, the evaluations can be reused to form other marginal contributions and thus save budget.
This idea can already be applied to the sampling of permutations of the player set.
\citet{Castro.2009} evaluate for each drawn permutation $\pi$ the marginal contribution $\Delta_i(\text{pre}_i(\pi))$ of each player $i$ to the preceding players in $\pi$.
Except for the last player in $\pi$, each evaluation $\nu(\text{pre}_i(\pi) \cup \{i\})$ can be reused for the marginal contribution of the succeeding player.

We further develop this paradigm of \emph{sample reusage} by exploiting the fact that any coalition $S \subseteq \mathcal{N}$ appears in $n$ many marginal contributions, one for each player, namely in $n-|S|$ many of the form $\Delta_i(S)$ for $i \notin S$ and $|S|$ many of the form $\Delta_i(S \setminus \{i\})$ for $i \in S$.
We meaningfully unify both cases by establishing the notion of an \emph{extended marginal contribution} in \cref{def:extended_mc}.
\begin{definition} \label{def:extended_mc}
    For any cooperative game $(\mathcal{N}, \nu)$, the extended marginal contribution of a player $i \in \mathcal{N}$ to a coalition $S \subseteq \mathcal{N}$ is given by
    \begin{equation*}
        \Delta'_i(S) := \nu(S \cup \{i\}) - \nu(S \setminus \{i\}) \, . 
    \end{equation*}
\end{definition}
Fittingly, this yields $\Delta_i'(S) = \Delta_i(S \setminus \{i\})$ for $i \in S$ and $\Delta_i'(S) = \Delta_i(S) = \Delta_i(S \setminus \{i\})$ for $i \notin S$.
Thus, we circumvent the case of $\Delta_i(S) = 0$ for $i \in S$.

We aim to draw in each round $m$ (of $M$ many) a coalition $S^{(m)} \subseteq \mathcal{N}$, compute the extended marginal contributions $\Delta'_i(S^{(m)})$ of \emph{all} players as illustrated in \cref{fig:lattice}, and update each $\hat\phi_i$ as the average of the corresponding observations:
\begin{equation} \label{eq:sample_mean2}
    \hat\phi_i = \frac{1}{M} \sum\limits_{m=1}^M \Delta'_i \left( S^{(m)} \right) \, .
\end{equation}
We reuse the coalition value $v_{S^{(m)}} = \nu(S^{(m)})$ to update all estimates by computing each extended marginal contribution as
\begin{equation}
    \Delta'_i \left( S^{(m)} \right) =  
    \begin{cases}
        v_{S^{(m)}} - \nu(S^{(m)} \setminus \{i\}) & \text{if } i \in S \\
        \nu(S^{(m)} \cup \{i\}) - v_{S^{(m)}} & \text{otherwise}
    \end{cases}
    \, .
\end{equation}
Consequently, updating all $n$ estimates requires only $n+1$ calls to $\nu$ such that we obtain a \emph{budget-efficiency} of $\frac{n}{n+1}$ sampled observations per call.
In comparison, drawing marginal contributions separately yields a budget-efficiency of $\nicefrac{1}{2}$.
In order to make this approach effective, it is desirable to obtain unbiased estimates leading to the question whether there even exists a probability distribution over $\mathcal{P}(\mathcal{N})$ to sample $S^{(m)}$ from such that $\mathbb{E}[\Delta'_i(S^{(m)})] = \phi_i$ for all $i \in \mathcal{N}$.
Indeed, we show its existence in \cref{pro:extended_mc} by means of a novel representation of the Shapley value based on extended marginal contributions.
\begin{proposition} \label{pro:extended_mc}
    For any cooperative game $(\mathcal{N}, \nu)$, the Shapley value of each player $i \in \mathcal{N}$ is a weighted average of its extended marginal contributions.
    In particular, it holds
    \begin{equation*}
        \phi_i = \sum\limits_{S \subseteq \mathcal{N}} \frac{1}{(n+1) \binom{n}{|S|}} \cdot \Delta'_i(S) \, .
    \end{equation*}
\end{proposition}
See \cref{app:CMCS_bias} for a proof.
The weighted average allows to view the Shapley value as the expected extended marginal contribution and thus drawing $S^{(m)}$ from the distribution
\begin{equation} \label{eq:CMCS_distribution}
    \mathbb{P} \left( S^{(m)}=S \right) = \frac{1}{(n+1) \binom{n}{|S|}} \hspace{0.2cm} \text{for all } S \subseteq \mathcal{N}
\end{equation}
yields unbiased estimates.
Note that this is indeed a well-defined probability distribution over $\mathcal{P}(\mathcal{N})$ as shown in \cref{app:CMCS_bias}.
The resulting algorithm \emph{Comparable Marginal Contributions Sampling} (CMCS) is given by \cref{alg:CMCS}.
It requires the cooperative game $(N,\nu)$, the budget $T$, and the parameter $k$ as input.
The number of performed rounds $M$ is bounded by $M = \lfloor \frac{T}{n+1} \rfloor$.
We solve sampling from the exponentially large power set of $\mathcal{N}$ by first drawing a size $\ell$ ranging from $0$ to $n$ uniformly at random (line 3) and then drawing uniformly a coalition $S$ of size $\ell$ (line 4).
This results in the probability distribution of \cref{eq:CMCS_distribution} since there are $n+1$ sizes and $\binom{n}{\ell}$ coalitions of size $\ell$ to choose from.
For the top-$k$ identification problem CMCS returns the set of $k$ many players $\hat{\mathcal{K}}$ for which it maintains the highest estimates $\hat\phi_i$.
Ties are solved arbitrarily.

\begin{algorithm}[ht]
    \caption{Comparable Marginal Contributions Sampling (CMCS)}
    \label{alg:CMCS}
    \textbf{Input}: $(\mathcal{N},\nu)$, $T \in \mathbb{N}, k \in [n]$
    \begin{algorithmic}[1]
        \STATE $\hat\phi_i \leftarrow 0$ for all $i \in \mathcal{N}$
        \FOR{$m = 1,\ldots,\lfloor\frac{T}{n+1}\rfloor$}
            \STATE Draw $\ell \in \{0,\ldots,n\}$ uniformly at random
            \STATE Draw $S \subseteq \mathcal{N}$ with $|S| = \ell$ uniformly at random
            \STATE $v_S \leftarrow \nu(S)$
            \FOR{$i \in \mathcal{N}$}
                \STATE $\Delta_i \leftarrow \begin{cases}
                    v_S - \nu(S \setminus \{i\}) & \text{if } i \in S \\
                    \nu(S \cup \{i\}) - v_S & \text{otherwise}
                \end{cases}$
                \STATE $\hat\phi_i \leftarrow \frac{(m-1) \cdot \hat\phi_i + \Delta_i}{m}$
            \ENDFOR
        \ENDFOR
    \end{algorithmic}
    \textbf{Output}: $\hat{\mathcal{K}}$ containing $k$ players with highest estimate $\hat\phi_i$
\end{algorithm}

\noindent
CMCS can also be applied for the approximate-all problem by simply returning its estimates since its sampling procedure and computation of estimates is independent of $k$.
Thus, it is also an unbiased equifrequent player-wise independent sampler (see \cref{sec:topk_vs_mse}) because the marginal contributions obtained in each round stem from a fixed joint distribution and the resulting marginal distributions coincide with \cref{eq:marginal_distribution} as implied by \cref{pro:extended_mc}.
Hence for $T$ being a multiple of $n+1$, its expected MSE is according to \cref{eq:MSE_result}:
\begin{equation} \label{eq:CMCS_MSE}
    \mathbb{E} [\text{MSE}] = \frac{n+1}{nT} \sum\limits_{i \in \mathcal{N}} \sigma_i^2 \, .
\end{equation}
For the top-$k$ identification the sampling scheme in CMCS yields an interesting property.
All players share extended marginal contributions to the same reference coalitions $S^{(m)}$.
Intuitively, this makes the estimates more comparable, as all have been updated using the same samples.
Instead of estimating $\phi_i$ and $\phi_j$ precisely, CMCS answers the relevant question whether $\phi_i > \phi_j$ holds, by comparing the players marginal contributions to roughly the same coalitions, modulo the case of $i \in S$ and $j \notin S$ or vice versa. 
Instead, drawing marginal contributions separately, independently between the players, can, metaphorically speaking, be viewed as comparing apples with oranges.

Consequently, the estimates $\hat\phi_i$ and $\hat\phi_j$ are correlated and we further conjecture that the
covariance $\text{Cov}(\Delta'_i(S^{(m)}), \Delta'_j(S^{(m)})) = \mathbb{E}[\Delta'_i(S^{(m))} \Delta'_j(S^{(m)})] - \mathbb{E}[\Delta'_i(S^{(m)})] \mathbb{E}[\Delta'_j(S^{(m)})]$ has a positive impact on the inclusion-exclusion error of CMCS in light of \cref{the:inc_exc_bound}.
For cooperative games in which the marginal contribution of a player is influenced by the coalitions size, our sampling scheme should yield positively correlated samples.
In this case, if player $i$ or $j$ is added to the same coalition $S$, it is likely that both have a positive marginal contribution (or both share a negative) which in turn speaks for a positive covariance.
For the general case, the covariance is stated in \cref{pro:covariance}.
\begin{proposition} \label{pro:covariance}
    For any cooperative game $(\mathcal{N}, \nu)$ the covariance between the extended marginal contributions of any players $i \neq j$ of the same round sampled by CMCS is given by
    \begin{multline*}
        \text{Cov} \left( \Delta'_i \left( S^{(m)} \right), \Delta'_j \left( S^{(m)} \right) \right) = \frac{1}{n+1} \sum\limits_{S \subseteq \mathcal{N} \setminus \{i\}} \Delta_i(S) \\
        \left( \frac{\Delta_j'(S)}{\binom{n}{|S|}} + \frac{\Delta'_j(S \cup \{i\})}{\binom{n}{|S|+1}} \right) - \phi_i \phi_j \, .
    \end{multline*}
\end{proposition}
The proof is given in \cref{app:CMCS_bias}.
The sum can be seen as the Shapley value $\phi_i$ in which each marginal contribution of $i$ is additionally weighted by extended marginal contributions of $j$.
To demonstrate the presumably positive covariance and give evidence to our conjecture, we consider a simple game of arbitrary size $n$ with $\nu(\mathcal{N}) = 1$ and $\nu(S)=0$ for all coalitions $S \neq \mathcal{N}$.
Each player has a Shapley value of $\frac{1}{n}$ and the covariance in \cref{pro:covariance} given by $\frac{1}{n+1} - \frac{1}{n^2}$ is strictly positive for $n \geq 2$.

\subsection{Relaxed Greedy Player Selection for Top-$k$ Identification}
\label{subsec:greedy}

Striving for budget-efficiency in the design of a sample procedure might be favorable, however, CMCS as proposed in \cref{subsec:CMCS} is forced to spend budget on the retrieval of marginal contributions for all players in order to maximize budget-efficiency.
This comes with the disadvantage that evaluations of $\nu$ are performed to sample for a player $i$ whose estimate $\hat\phi_i$ is possibly already reliable enough and does not need further updates compared to other players.
This does not even require $\hat\phi_i$ to be precise in absolute terms.
Instead, it is sufficient to predict with certainty whether $i$ belongs to the top-$k$ or not by comparing it to the other estimates.
This observation calls for a more selective mechanism deciding which players to leave out in each round and thus save budget.

A radical approach is the greedy selection of a single player which maximizes a \emph{selection criterion} based on the collected observations that incorporates incentives for exploration and exploitation.
Gap-E \cite{Gabillon.2011,Bubeck.2013} composes the selection criterion out of the uncertainty of a player's top-$k$ (exploitation) membership and its number of observations (exploration).
Similarly, BUS \cite{Kolpaczki.2021} selects the player $i$ minimizing the product of its estimate's distance to the predicted top-$k$ border $\frac{1}{2} (\min_{i \in \hat{\mathcal{K}}} \hat\phi_i - \max_{j \in \mathcal{N} \setminus \hat{\mathcal{K}}} \hat\phi_j)$ times its sample number $M_i$.
In the same spirit but outside of the fixed-budget setting, SHAP$@$K \cite{Kariyappa.2024} chooses for given $\delta \in (0,1)$ the two players $i \in \hat{\mathcal{K}}$ and $j \in \mathcal{N} \setminus \hat{\mathcal{K}}$ with the highest overlap in their $\nicefrac{\delta}{n}$-confidence intervals of their estimates $\hat\phi_i$ and $\hat\phi_j$.
It applies a stopping condition and terminates when no overlaps between $\hat{\mathcal{K}}$ and $\mathcal{N} \setminus \hat{\mathcal{K}}$ larger then a specified error $\varepsilon$ exist.
Assuming normally distributed estimates $\hat\phi_i$ under the central limit theorem, it holds $\mathbb{P}(\rho_{\text{inc+exc}}(\hat{\mathcal{K}}) \leq \varepsilon) \geq 1 - \delta$ for its prediction $\mathcal{\hat{K}}$.

\begin{figure*}[ht]
    \begin{center}
        \includegraphics[width=0.95\textwidth]{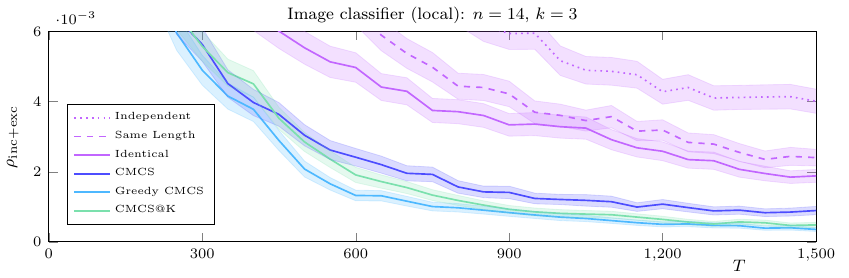}
        \vskip -0.15in
        \caption{
            Inclusion-exclusion error $\varepsilon$ for increasingly comparable sampling variants (\emph{Independent}, \emph{Same Length}, \emph{Identical}), incorporation of sample-reusage (CMCS), and greedy selection (Greedy CMCS, CMCS$@$K) depending on $T$.
        }
        \label{fig:CMCS_results}
    \end{center}
    \vskip -0.15in
\end{figure*}

Given the core idea of CMCS to draw samples for multiple players at once in order to increase budget-efficiency and obtain correlated observations, the greedy selection of a single player as done in \cite{Gabillon.2011,Kolpaczki.2021} or just a pair \cite{Kariyappa.2024} is not suitable for our method.
The phase-wise elimination performed by SAR \cite{Bubeck.2013} is not viable as it assumes all observations to be independent in order to analytically derive phase lengths.
Instead, we relax the greediness by probabilistically selecting a set of players $P^{(m)} \subseteq \mathcal{N}$ in each round $m$, favoring those players who fulfill a selection criterion to higher degree.
By doing so, we propose \emph{Greedy CMCS} that intertwines the overcoming of the exploration-exploitation dilemma with the pursuit of budget-efficiency.
We do not abandon exploration, since every player gets a chance to be picked, and the selection criterion incentivizes exploitation as it reflects how much the choice of a player benefits the prediction $\hat{\mathcal{K}}$.

Our selection criterion is based on the current knowledge of $\hat\phi_1,\ldots,\hat\phi_n$ and the presumably best players $\hat{\mathcal{K}}$.
Inspired by \cref{the:inc_exc_bound}, we approximate the probability of each pair of players $i \in \hat{\mathcal{K}}$ and $j \in \mathcal{N} \setminus \hat{\mathcal{K}}$ being incorrectly partitioned by Greedy CMCS as
\begin{equation} \label{eq:likelihood}
    \hat{p}_{i,j} := \Phi \left( \sqrt{M_{i,j}} \frac{\hat\delta_{i,j}}{\hat\sigma_{i,j}} \right) \, .
\end{equation}
For all pairs $(i,j) \in \mathcal{N}^2$ we track:
\begin{itemize}
    \item the number of times $M_{i,j}$ that both $i$ and $j$ have been selected in a round,
    \item the sampled marginal contributions' mean difference $\hat\delta_{i,j} := \frac{1}{M_{i,j}} \sum\limits_{m=1}^{M_{i,j}} \Delta'_j(S^{(f_{i,j}(m))}) - \Delta'_i(S^{(f_{i,j}(m))})$ within these $M_{i,j}$ rounds, where $f_{i,j}(m)$ denotes the $m$-th round in which $i$ and $j$ are selected, and
    \item the estimate $\hat\sigma_{i,j}^2$ of the variance $\sigma_{i,j}^2 := \mathbb{V}[\Delta'_i(S^{(m)}) - \Delta'_j(S^{(m)})]$ w.r.t.\ \cref{eq:CMCS_distribution}.
\end{itemize}

\noindent
Important to note is that we may not simply use the difference $\hat\phi_j - \hat\phi_i$ of our Shapley estimates, including all rounds, instead of $\hat{\delta}_{i,j}$ because $\hat\phi_i$ and $\hat\phi_j$ may differ in their respective total amount of total samples $M_i$ and $M_j$ such that the central limit theorem used for \cref{the:inc_exc_bound} is not applicable anymore.
We derive \cref{eq:likelihood} in \cref{app:pairwise_prob}.

\begin{figure*}[ht]
    \begin{center}
        \includegraphics[width=0.95\textwidth]{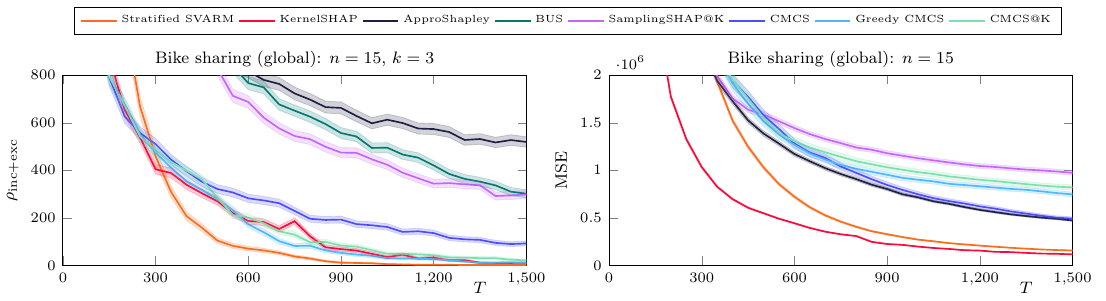}
        \vskip -0.15in
        \caption{
            Comparison of achieved inclusion-exclusion error of various algorithms for top-$k$ identification (left) and approximate-all (right) depending on $T$.
        }
        \label{fig:topk_vs_mse}
    \end{center}
    \vskip -0.15in
\end{figure*}

For each pair $(i,j) \in \hat{\mathcal{K}} \times (\mathcal{N} \setminus \hat{\mathcal{K}})$ the estimate $\hat{p}_{i,j}$ quantifies how likely $i$ and $j$ are wrongly partitioned: Greedy CMCS estimates $\hat\phi_i \geq \hat\phi_j$ although $\phi_i < \phi_j$ holds.
Since we want to minimize the probability of such a mistake, it comes natural to include the pair $(i,j)$ with the highest estimate $\hat{p}_{i,j}$ in the next round of Greedy CMCS to draw marginal contributions from, i.e.\ $i,j \in P^{(m)}$.
As a consequence, $\hat\phi_i$ and $\hat\phi_j$ should become more reliable causing the error probability to shrink.
Let $Q^{(m)} \subseteq \hat{\mathcal{K}} \times (\mathcal{N} \setminus \hat{\mathcal{K}})$ be the set of selected pairs in round $m$ from which the selected players are formed as $P^{(m)} = \{i \in \mathcal{N} \mid \exists (i,j) \in Q^{(m)} \lor \exists (j,i) \in Q^{(m)} \}$.
In order to allow for more than two updated players in a round $m$, i.e.\ $|Q^{(m)}| > 1$ , but waive pairs that are more likely to be correctly classified, we probabilistically include pairs in $Q^{(m)}$ depending on their $\hat{p}$-value.
Let $\hat{p}_\text{max} = \max_{i \in \hat{\mathcal{K}}, j \notin \hat{\mathcal{K}}} \hat{p}_{i,j}$ be the currently highest and $\hat{p}_\text{min} = \min_{i \in \hat{\mathcal{K}}, j \notin \hat{\mathcal{K}}} \hat{p}_{i,j}$ the currently lowest value.
We select each pair $(i,j)$ independently with probability
\begin{equation} \label{eq:greedy_selection}
   \mathbb{P} \left( (i,j) \in Q^{(m)} \right) = \frac{\hat{p}_{i,j} - \hat{p}_\text{min}}{\hat{p}_\text{max}-\hat{p}_\text{min}} \hspace{0.2cm} \text{for all } (i,j) \in \hat{\mathcal{K}} \times (\mathcal{N} \setminus \hat{\mathcal{K}}) \, .
\end{equation}
This forces the pair with $\hat{p}_{\text{max}}$ to be picked and that with $\hat{p}_{\text{min}}$ to be left out.
The probability of a pair beings elected increases monotonically with its $\hat{p}$-value.

Within an executed round we do not only collect marginal contributions for players in $P^{(m)}$ and update $M_{i,j}$, $\hat\delta_{i,j}$, and $\hat\sigma_{i,j}^2$ for all $(i,j) \in Q^{(m)}$.
We use the collected information to its fullest by also updating the estimates of all pairs $(i,j)$ with both players being present in $P^{(m)}$ despite $(i,j) \notin Q^{(m)}$.
Visually speaking, we update the complete subgraph induced by $P^{(m)}$ with players being nodes and edges containing the pairwise estimates.

Since the assumption of normally distributed estimates motivated by the central limit theorem is not appropriate for a low number of samples, we initialize Greedy CMCS with a warm-up phase as proposed for SHAP@K \cite{Kariyappa.2024}.
During the warm-up $M_{\text{min}}$ many rounds of CMCS are performed such that afterwards every player's Shapley estimate is based on $M_{\text{min}}$ samples.
This consumes a budget of $(n+1) M_{\text{min}}$ many evaluations.
$M_{\min}$ is provided to Greedy CMCS as a parameter.
Subsequently, the above described round-wise greedy sampling is applied as the second phase until the depletion of the in total available budget $T$.
The pseudocode of the resulting algorithm Greedy CMCS is given in \cref{app:pseudocode}.

Instead of our proposed selection mechanism, one can sample in the second phase only from the two players $i \in \hat{\mathcal{K}}$ and $j \notin \hat{\mathcal{K}}$ with the biggest overlap in confidence intervals as performed by SHAP@K.
Leaving the sampling of CMCS in the first phase untouched, we call this variant \emph{CMCS$@$K}.
This is feasible since the choice of the sampling procedure in SHAP$@$K is to some extent arbitrary, as long as it yields confidence intervals for the Shapley estimates.

\section{Empirical Results}
\label{sec:emprical_results}

We conduct multiple experiments of different designs to assess the performance of sampling comparable marginal contributions at the example of explanation tasks on real-world datasets.
First, we demonstrate in \cref{subsec:CMCS_results} the iterative improvements of our proposed algorithmic tricks ranging from the naive independent sampling to Greedy CMCS and CMCS$@$K.
\cref{subsec:mse_vs_eie} investigates whether favorable MSE values of algorithms for the approximate-all problem translate on the same cooperative games to the inclusion-exclusion error for top-$k$ identification.
In \cref{subsec:SOTA_comparison} we compare our variants of CMCS against baselines and state-of-the-art competitors.
Lastly, we investigate in \cref{subsec:PAC} the required budget until the stopping criterion of \cite{Kariyappa.2024} applied to CMCS guarantees an error of at most $\varepsilon$ with probability at least $1-\delta$.
\begin{figure*}[ht]
    \begin{center}
        \includegraphics[width=0.95\textwidth]{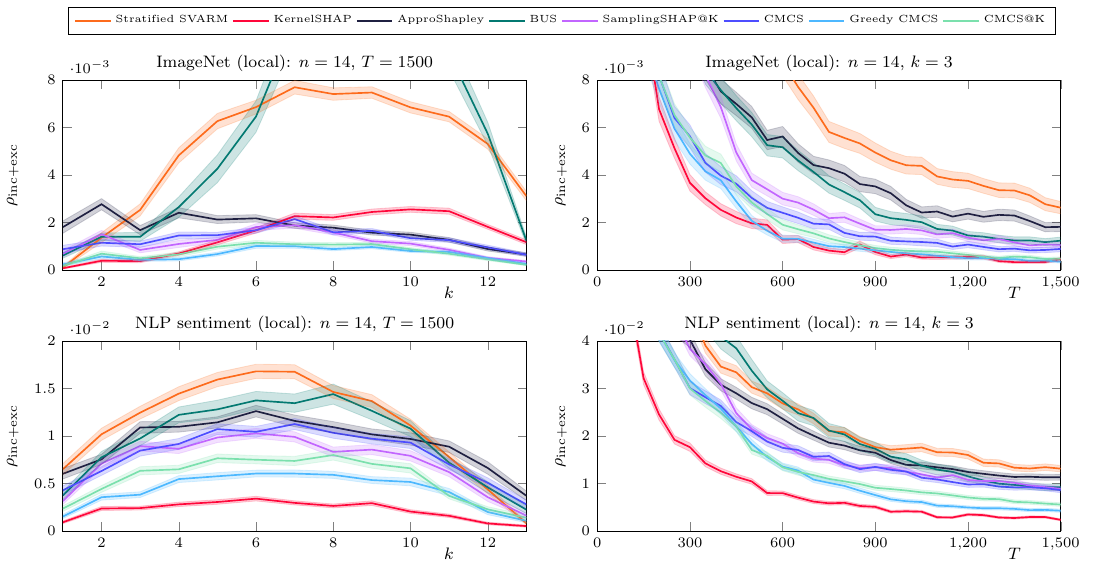}
        \vskip -0.15in
        \caption{
            Comparison of achieved inclusion-exclusion error with baselines for local explanations: fixed budget with varying $k$ (left) and fixed $k$ with increasing budget (right).
        }
        \label{fig:sota_local}
    \end{center}
    \vskip -0.15in
\end{figure*}
All performance measures are calculated by exhaustively computing the Shapley values in advance and averaging the results over 1000 runs.
Standard errors are included as shaded bands.
We compare against \emph{ApproShapley} \cite{Castro.2009}, \emph{KernelSHAP} \cite{Lundberg.2017} (with reference implementation provided by the \verb|shap| python package, the one to sample without replacement), \emph{Stratified SVARM} \cite{Kolpaczki.2024a}, \emph{BUS} \cite{Kolpaczki.2021}, and \emph{SamplingSHAP$@$K} \cite{Kariyappa.2024} which is SHAP@K drawing samples according to ApproShapley.
For both SamplingSHAP$@$K and CMCS$@$K, we use $M_\text{min}=30$ and confidence intervals of $\nicefrac{\delta}{n}$ with $\delta=0.001$.
We drop \emph{Gap-E} \cite{Gabillon.2011} and \emph{SAR} \cite{Bubeck.2013} due to worse performances
\footnote{All code can be found at \url{https://github.com/timnielen/top-k-shapley}}.

\paragraph{Datasets and games.}
Analogously to \cite{Kolpaczki.2024a,Kolpaczki.2024b}, we generate cooperative games from two types of explanation tasks in which the Shapley values represent feature importance scores.
For global games, we construct the value function by training a \emph{sklearn} random forest with 20 trees on each feature subset and taking its classification accuracy, or the $R^2$-metric for regression tasks, on a test set as the coalitions' worth.
We employ the \emph{Adult} ($n=14$, classification), \emph{Bank Marketing} ($n=16$, classification), \emph{Bike Sharing} ($n=15$, regression), \emph{Diabetes} ($n=10$, regression), \emph{German Credit} ($n=20$, classification),  \emph{Titanic} ($n=11$, classification), and \emph{Wine} ($n=13$, classification) dataset. 
For local games, we create a game by picking a random datapoint and taking a pretrained model's prediction value as each coalition's worth.
Feature values are imputed by their mean, respectively mode.
For this purpose we take the \emph{Adult} ($n=14$, XGBoost, classification), \emph{ImageNet} ($n=14$, ResNet18, classification), and \emph{NLP Sentiment} ($n=14$, DistilBERT transformer, regression, IMDB data)~dataset. 

\subsection{Advantage of Comparable Sampling}
\label{subsec:CMCS_results}

Greedy CMCS builds upon multiple ideas whose effects onto the approximation quality is depicted  in isolation by \cref{fig:CMCS_results}.
As a baseline we consider the \emph{independent} sampling of marginal contributions of each player with distribution given in \cref{eq:marginal_distribution}.
The comparability of the samples is stepwise increased by sampling in each round marginal contributions to coalitions of the \emph{same length} for all players, and next using the \emph{identical} coalition $S^{(m)}$ drawn according to \cref{eq:CMCS_distribution}.
In compliance with our conjecture, the decreasing error from \emph{independent} to \emph{same length} and further to \emph{identical} speaks in favor of the beneficial impact that comes with correlated observations.
The biggest leap in performance is caused by reusing the evaluated worth $\nu(S^{(m)})$ appearing in each marginal contribution of the \emph{independent} variant resulting in CMCS.
The sample reusage alone almost doubles the budget-efficiency from $\nicefrac{1}{2}$ to $\nicefrac{n}{n+1}$.
On top of that, incorporating (relaxed) greedy sampling gifts Greedy CMCS and CMCS@K a further advantage by halving the error for higher budget ranges.

\subsection{MSE vs.\ Inclusion-Exclusion Error}
\label{subsec:mse_vs_eie}

Given the similarities between the problem statements of approximating all Shapley values (cf.\ \cref{subsec:approximate_all}) and that of top-$k$ identification (cf.\ \cref{subsec:top_k_identification}) at first sight, one might suspect that approximation algorithms performing well in the former, also do so in the latter and vice versa.
However, \cref{fig:topk_vs_mse} shows a different picture.
The best performing methods Stratified SVARM and KernelSHAP remain consistent but change in order.
The variants of CMCS are less favorable in terms of MSE but are barely outperformed in top-$k$ identification.
We interpret this as further evidence that top-$k$ identification indeed rewards positively correlated samples supporting our intuition of comparability.
Most striking is the difference between ApproShapley and CMCS.
Assuming to know $\nu(\emptyset)=0$, ApproShapley exhibits a budget-efficiency of $1$ as it consumes in each sampled permutation $n$ evaluations and retrieves $n$ marginal contributions, which is only slightly better than that of CMCS with $\nicefrac{n}{n+1}$.
Thus, it should be only marginally better in approximation according to \cref{eq:MSE_result} and \cref{eq:CMCS_MSE}.
Our results in \cref{fig:topk_vs_mse} confirm the precision of our theory.
However, notice how CMCS significantly outperforms ApproShapley in terms of $\rho_{\text{inc+exc}}$ despite the almost identical budget usage.
Hence, it is the stronger correlation of samples drawn by CMCS combined with the nature of top-$k$ identification that causes the observed advantage of comparable sampling.

\subsection{Comparison with Existing Methods}
\label{subsec:SOTA_comparison}
\begin{figure*}[ht]
    \begin{center}
        \includegraphics[width=0.95\textwidth]{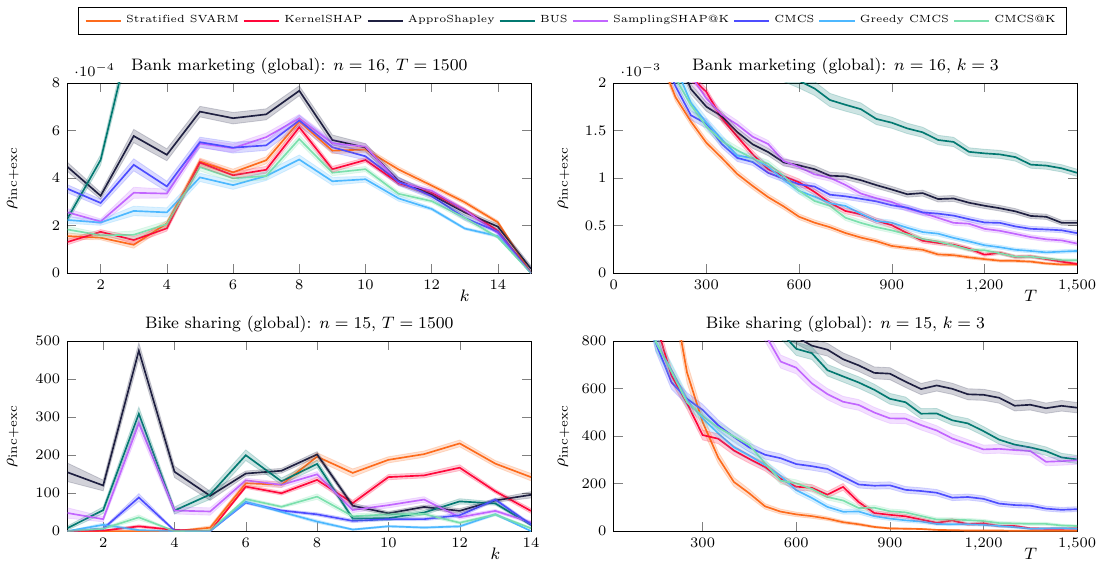}
        \vskip -0.15in
        \caption{
            Comparison of achieved inclusion-exclusion error with baselines for global explanations: fixed budget with varying $k$ (left) and fixed $k$ with increasing budget (right).
        }
        \label{fig:sota_global}
    \end{center}
    \vskip -0.15in
\end{figure*}
\cref{fig:sota_local} and \ref{fig:sota_global} compare the performances of our methods against baselines for local and global games.
For fixed $k=3$,
we observe the competitiveness of Greedy CMCS and CMCS$@$K being mostly on par with KernelSHAP, but getting beaten by Stratified SVARM for global games, which in turns subsides at local games.
Greedy CMCS exhibits stable performance across both explanation types and the whole range of $k$.
On the other hand, if instead the budget is fixed, Greedy CMCS has often the upper hand for values of $k$ close to $\nicefrac{n}{2}$ and is even with KernelSHAP for lower $k$.

\subsection{Budget Consumption for PAC Solution}
\label{subsec:PAC}

Assuming normally distributed Shapley estimates, SHAP$@$K is a $(\varepsilon,\delta)$-PAC learner \cite{Kariyappa.2024}, i.e.\ upon self-induced termination it holds $\rho_{\text{inc+exc}}(\hat{\mathcal{K}}) \leq \varepsilon$ with probability at least $1-\delta$.
KernelSHAP is not~applicable as it does not yield confidence bounds.
For this reason \citet{Kariyappa.2024} sample marginal contributions referred as \emph{SamplingSHAP$@$K}.
Its stopping condition is triggered as soon as no $\nicefrac{\delta}{n}$ confidence intervals for the estimates $\hat\phi_i$ overlap between $\hat{\mathcal{K}}$ and $\mathcal{N} \setminus \hat{\mathcal{K}}$.
We apply~the stopping condition to our algorithms and compare to SamplingSHAP$@$K in the PAC-setting.
\cref{tab:pac_results} shows~the average number of calls to $\nu$ until termination that is~to~be minimized.
For some local games the number of calls~is significantly higher due to the large variance in the difficulty of the respective games induced from each datapoint.
CMCS$@$K shows the best results in nearly every game~by some margin, which makes it the algorithm of~choice~for PAC-learning.
Thus, CMCS@K is preferable when guarantees for approximation quality are required and improves upon SHAP$@$K due to its refined sampling~mechanism.
\begin{table*}[h]
    \centering
    \footnotesize
    \renewcommand{\arraystretch}{1.2}
    \begin{tabular}{lcrrrrrrrr}
        & & \multicolumn{2}{c}{SamplingSHAP@K} & \multicolumn{2}{c}{CMCS} & \multicolumn{2}{c}{CMCS@K} & \multicolumn{2}{c}{Greedy CMCS} \\
        \cmidrule(r){3-4} \cmidrule(r){5-6} \cmidrule(r){7-8} \cmidrule(r){9-10}
        Game & $n$ & \#samples & SE & \#samples & SE & \#samples & SE & \#samples & SE \\
        \hline
        Adult (global) & 14 & 38\,998 & 1\,247 & 137\,861 & 2\,517 & \textbf{30\,995} & 673 & 39\,071 & 738 \\
        German credit (global) & 20 & 21\,939 & 336 & 56\,738 & 1\,129 & \textbf{16\,437} & 248 & 22\,327 & 328 \\
        Bike sharing (global) & 15 & 4\,850 & 97 & 13\,053 & 164 & \textbf{3\,982} & 54 & 8\,894 & 117 \\
        Bank marketing (global) & 16 & 15\,124 & 287 & 39\,144 & 875 & \textbf{12\,000} & 206 & 16\,260 & 267 \\
        Diabetes (global) & 10 & 3\,723 & 94 & 7\,793 & 143 & \textbf{2\,976} & 55 & 4\,593 & 85 \\
        Titanic (global) & 11 & 4\,852 & 113 & 11\,036 & 237 & \textbf{3\,884} & 72 & 5\,782 & 124 \\
        Wine (global) & 13 & 34\,953 & 1\,046 & 120\,859 & 1\,906 & \textbf{29\,913} & 641 & 34\,265 & 501 \\
        NLP sentiment (local) & 14 & 626\,346 & 188\,125 & 3\,351\,274 & 764\,663 & 568\,261 & 156\,674 & \textbf{447\,252} & 77\,149 \\
        ImageNet (local) & 14 & 135\,851 & 39\,335 & 578\,670 & 196\,181 & \textbf{108\,267} & 32\,067 & 261\,586 & 147\,126 \\
        Adult (local) & 14 & 18\,464 & 4\,391 & 55\,779 & 17\,954 & \textbf{14\,406} & 3\,645 & 16\,160 & 3\,765 \\
    \end{tabular}
    \caption{Average number of calls to $\nu$ in the PAC-setting (see \cref{eq:PAC}) across different datasets averaged over 200 runs using $\delta=0.01$ and $\epsilon=0.0005$ for $k=5$.}
    \label{tab:pac_results}
\end{table*}

\section{Conclusion}
\label{sec:conclusion}

We emphasized differences between the problem of approximating all Shapley values and that of identifying the $k$ players with highest Shapley values.
Analytically recognizing the advantage that correlated samples promise, we developed with CMCS an antithetic sampling algorithm that reuses evaluations to save budget.
Our extensions Greedy CMCS and CMCS$@$K employ selective strategies for sampling.
Both demonstrate competitive performances, with Greedy CMCS being better suited for fixed budgets, whereas CMCS$@$K is clearly favorable in the PAC-setting.
Our proposed methods are not only model-agnostic, moreover, they can handle any cooperative game, facilitating their application for any explanation type and domain even outside of explainable AI.
The difficulties that some algorithms face when translating their performance to top-$k$ identification suggest that practitioner's being consciously interested in top-$k$ explanations might have an advantage by applying tailored top-$k$ algorithms instead of the trivial reduction to the approximate-all problem.
Future work could investigate the sensible choice of the warm-up length in Greedy CMCS and CMCS$@$K which poses a trade-off between exploration and exploitation.
Modifying our considered problem statement to identify the players with highest absolute Shapley values poses an intriguing variation for detecting the most impactful players and opens the door to new approaches.
Finally, Shapley interactions enrich Shapley-based explanations.
The number of pairwise interactions grows quadratically with $n$, hence top-$k$ identification could play an even more significant role.
Our work can be understood as a methodological precursor to such extensions.

\bibliography{references}
\bibliographystyle{icml2025}

\newpage
\appendix
\onecolumn

\section{Theoretical Analysis}
\label{app:analysis}

\subsection{Proof of \cref{the:inc_exc_bound}}
\label{app:lower_bound}

For the estimate $\hat{\mathcal{K}} \subseteq \mathcal{N}$ returned by an algorithm for the top-$k$ identification problem we can obviously state
\begin{equation*}
	\begin{array}{l}
    \mathbb{P} (\hat{\mathcal{K}} \in \mathbb{K}_{\varepsilon}) = \sum\limits_{\mathcal{K} \in \mathbb{K}_{\varepsilon}} \mathbb{P}(\hat{\mathcal{K}} = \mathcal{K}) \, .
	\end{array}
\end{equation*}
Given the construction of $\hat{\mathcal{K}}$, $\mathcal{A}$ must choose any $i \in \mathcal{N}$ to be in $\hat{\mathcal{K}}$ if $\hat\phi_i > \hat\phi_j$ holds for at least $n-k$ many players $j \in \mathcal{N}$.
Hence, for any $\mathcal{K} \in \mathbb{K}_\varepsilon$ we have:
\begin{equation*}
	\begin{array}{rl}
    \mathbb{P} (\hat{\mathcal{K}} = \mathcal{K}) \geq & \ \mathbb{P} (\forall i \in \mathcal{K} \ \forall j \in \mathcal{N} \setminus \mathcal{K} : \hat\phi_i > \hat\phi_j) \\
    = & \ 1 - \mathbb{P} (\exists i \in \mathcal{K} \ \exists j \in \mathcal{N} \setminus \mathcal{K} : \hat\phi_i \leq \hat\phi_j) \\
    \geq & \ 1 - \sum\limits_{\substack{i \in \mathcal{K} \\ j \in \mathcal{N} \setminus \mathcal{K}}} \mathbb{P} (\hat\phi_i \le \hat\phi_j)
	\end{array}
\end{equation*}

Given the assumptions on the sampling procedure and the aggregation to estimates $\hat\phi_1,\ldots,\hat\phi_n$, we can apply the central limit theorem (CLT) to state that for any $i \in \mathcal{K}$ and $j \in \mathcal{N} \setminus \mathcal{K}$ the distribution of
$\sqrt{M} \left( (\hat\phi_i - \hat\phi_j) - (\phi_i - \phi_j) \right)$
converges to a normal distribution with mean $0$ and variance $\sigma_{i,j}^2$ as $M \to \infty$ since $\mathbb{E}[\hat\phi_i - \hat\phi_j] = \phi_i - \phi_j$.
Although $M$ is finite as it is limited by the budget $T$, we assume it to be normally distributed, to which it comes close to in practice for large $M$.
Hence, for any $i \in \mathcal{K}$ and $j \in \mathcal{N} \setminus \mathcal{K}$ we derive:
\begin{equation*}
	\begin{array}{rl}
    \mathbb{P} (\hat\phi_i \leq \hat\phi_j) = & \ \mathbb{P} (\hat\phi_i -\hat\phi_j \leq 0) \\
    = & \ \mathbb{P} ((\hat\phi_i - \hat\phi_j) - (\phi_i - \phi_j) \leq - (\phi_i - \phi_j)) \\
    = & \ \mathbb{P} (\sqrt{M} ((\hat\phi_i -\hat\phi_j) - (\phi_i -\phi_j)) \leq \sqrt{M}(\phi_j - \phi_i)) \\
    \stackrel{CLT}{=} & \ \Phi \left( \sqrt{M} \frac{\phi_j - \phi_i}{\sigma_{i,j}} \right)
	\end{array}
\end{equation*}
where $\Phi$ is the standard normal cumulative distribution function.
Putting the intermediate results together, we obtain
\begin{equation*}
	\begin{array}{l}
    \mathbb{P} (\hat{\mathcal{K}} \in \mathbb{K}_{\varepsilon}) \geq \sum\limits_{\mathcal{K} \in \mathbb{K}_\epsilon} \left[ 1-\sum\limits_{\substack{i \in \mathcal{K} \\ j \in \mathcal{N} \setminus \mathcal{K}}} \Phi \left( \sqrt{M} \frac{\phi_j -\phi_i}{\sigma_{i,j}} \right) \right] \, .
	\end{array}
\end{equation*}

\subsection{Comparable Marginal Contributions Sampling}
\label{app:CMCS_bias}

Proof that \cref{eq:CMCS_distribution} induces a well-defined probability distribution: \\
Obviously it holds $\mathbb{P}(S) \geq 0$ and for the sum of probabilities we have:
\begin{equation*}
	\begin{array}{l}
    \sum\limits_{S \subseteq \mathcal{N}} \mathbb{P}(S)
    = \sum\limits_{S \subseteq \mathcal{N}} \frac{1}{(n+1)\binom{n}{|S|}}
    = \sum\limits_{\ell = 0}^{n} \sum\limits_{\substack{S \subseteq \mathcal{N} \\ |S| = l}} \frac{1}{(n+1)\binom{n}{\ell}}
    = \sum\limits_{\ell = 0}^{n} \frac{\binom{n}{\ell}}{(n+1)\binom{n}{\ell}}
    =  1 \, .
	\end{array}
\end{equation*}
\textbf{Proof of \cref{pro:extended_mc}:} \\
For any $i \in \mathcal{N}$ we derive:
\begin{equation*}
	\begin{array}{rl}
    \sum\limits_{S \subseteq \mathcal{N}} \frac{1}{(n+1)\binom{n}{|S|}} \cdot \Delta'_i(S) = & \ \sum\limits_{\substack{S \subseteq \mathcal{N} \\ i \in S}} \frac{1}{(n+1)\binom{n}{|S|}} \cdot \Delta_i(S \setminus \{i\}) + \sum\limits_{\substack{S \subseteq \mathcal{N} \\ i \notin S}} \frac{1}{(n+1)\binom{n}{|S|}} \cdot \Delta_i(S) \\
    = & \ \sum\limits_{S \subseteq \mathcal{N} \setminus \{i\}} \frac{1}{(n+1)\binom{n}{|S|+1}} \cdot \Delta_i(S) + \sum\limits_{S \subseteq \mathcal{N} \setminus \{i\}} \frac{1}{(n+1)\binom{n}{|S|}} \cdot \Delta_i(S) \\
    = & \ \sum\limits_{S \subseteq \mathcal{N} \setminus \{i\}} \frac{1}{n+1} \left( \frac{1}{\binom{n}{|S|+1}} + \frac{1}{\binom{n}{|S|}} \right) \cdot \Delta_i(S) \\
    = & \ \sum\limits_{S \subseteq \mathcal{N} \setminus \{i\}} \frac{1}{n \cdot \binom{n-1}{|S|}} \cdot \Delta_i(S) \\
    = & \ \phi_i
	\end{array}
\end{equation*}
\noindent
\textbf{Proof of \cref{pro:covariance}:} \\
Given the unbiasedness of the samples, i.e. $\mathbb{E}[\Delta'_i(S^{(m)})] = \phi_i$ for every $i \in \mathcal{N}$, the covariance is given by:
\begin{equation*}
	\begin{array}{rl}
    \text{Cov} \left( \Delta'_i(S^{(m)}), \Delta'_j(S^{(m)}) \right) = & \ \mathbb{E} \left[ \Delta'_i(S^{(m)}) \Delta'_j(S^{(m)}) \right] - \mathbb{E} \left[ \Delta'_i(S^{(m)}) \right] \mathbb{E} \left[ \Delta'_j(S^{(m)}) \right] \\
    = & \ \mathbb{E} \left[ \Delta'_i(S^{(m)}) \Delta'_j(S^{(m)}) \right] - \phi_i \phi_j \\
	\end{array}
\end{equation*}
For the first term we derive:
\begin{equation*}
	\begin{array}{rl}
    & \ \mathbb{E} \left[ \Delta'_i(S^{(m)}) \Delta'_j(S^{(m)}) \right] \\
    = & \ \sum\limits_{S \subseteq \mathcal{N}} \frac{1}{(n+1) \binom{n}{|S|}} \cdot \Delta'_i(S) \Delta'_j(S) \\
    = & \ \frac{1}{n+1} \sum\limits_{S \subseteq \mathcal{N} \setminus \{i,j\}} \frac{\Delta_i(S) \Delta_j(S)}{\binom{n}{|S|}} + \frac{\Delta_i(S) \Delta_j(S \cup \{i\})}{\binom{n}{|S|+1}} + \frac{\Delta_i(S \cup \{j\}) \Delta_j(S)}{\binom{n}{|S|+1}} + \frac{\Delta_i(S \cup \{j\}) \Delta_j(S \cup \{i\})}{\binom{n}{|S|+2}} \\
    = & \ \frac{1}{n+1}  \sum\limits_{S \subseteq \mathcal{N} \setminus \{i,j\}} \Delta_i(S) \cdot \left( \frac{\Delta_j(S)}{\binom{n}{|S|}} + \frac{\Delta_j(S \cup \{i\})}{\binom{n}{|S|+1}}\right) + \Delta_i(S \cup \{j\}) \cdot \left( \frac{\Delta_j(S)}{\binom{n}{|S|+1}} + \frac{\Delta_j(S \cup \{i\})}{\binom{n}{|S|+2}} \right) \\
    = & \ \frac{1}{n+1} \sum\limits_{S \subseteq \mathcal{N} \setminus \{i\}} \Delta_i(S) \cdot \left( \frac{\Delta_j'(S)}{\binom{n}{|S|}} + \frac{\Delta'_j(S \cup \{i\})}{\binom{n}{|S|+1}} \right)
	\end{array}
\end{equation*}

\subsection{Approximating Pairwise Probabilities for Greedy CMCS}
\label{app:pairwise_prob}

Analogously to \cref{app:lower_bound}, we derive for any pair $i,j \in \mathcal{N}$ and unbiased equifrequent player-wise independent sampler:
\begin{equation*}
	\begin{array}{rl}
    \mathbb{P}(\phi_i < \phi_j) = & \ \mathbb{P}(\phi_i - \phi_j < 0) \\
    = & \ \mathbb{P}((\hat\phi_i - \hat\phi_j) - (\phi_i - \phi_j) > \hat\phi_i - \hat\phi_j) \\
    = & \ \mathbb{P}(\sqrt{M}((\hat\phi_i -\hat\phi_j) - (\phi_i - \phi_j)) > \sqrt{M}(\hat\phi_i -\hat\phi_j)) \\
    \stackrel{CLT}{=} & \ \Phi \left( \sqrt{M} \frac{\hat\phi_j - \hat\phi_i}{\sigma_{i,j}} \right)
	\end{array}
\end{equation*}
Since this statement does not require the knowledge of an eligible coalition $\mathcal{K}$, we can estimate the likelihood of $\phi_i < \phi_j$ during runtime of the approximation algorithm.
For this purpose, we use the sample variance to estimate $\sigma_{i,j}$.
Note that $M$ is the number of drawn samples that both $\hat\phi_i$ and $\hat\phi_j$ share.
Since the players' marginal contributions are selectively sampled, Greedy CMCS substitutes $M$ by the true number of joint appearances $M_{i,j}$ and $\hat\phi_i - \hat\phi_j$ by $\hat\delta_{i,j}$ which only takes into account marginal contributions of $i$ and $j$ which have been acquired during rounds in which both players have been selected.

\newpage
\section{Pseudocode of Greedy CMCS}
\label{app:pseudocode}

In addition to the pseudocode in \cref{alg:GreedyCMCS}, we provide further details regarding the tracking of estimates and probabilistic selection of players.
\begin{algorithm}[H]
    \caption{Greedy CMCS}
    \label{alg:GreedyCMCS}
    \textbf{Input}: $(\mathcal{N},\nu)$, $T \in \mathbb{N}, k \in [n], M_\text{min}$
    \begin{algorithmic}[1]
        \STATE $\hat\phi_i, M_i \leftarrow 0$ for all $i \in \mathcal{N}$
        \STATE $M_{i,j}, \Sigma_{i,j}, \Gamma_{i,j} \leftarrow 0$ for all $i,j \in \mathcal{N}$
        \STATE $t \leftarrow 0$
        \WHILE{$t < T$}
            \STATE Draw $\ell \in \{0,\ldots,n\}$ uniformly at random
            \STATE Draw $S \subseteq \mathcal{N}$ with $|S| = l$ uniformly at random
            \STATE $v_S \leftarrow \nu(S)$
            \STATE $t \leftarrow t+1$
            \STATE $P \leftarrow \textsc{\texttt{SelectPlayers}}$
            \FOR{$i \in P$}
                \IF{$t = T$}
                    \STATE exit
                \ENDIF
                \STATE $\Delta_i \leftarrow \begin{cases}
                    v_S - \nu(S \setminus \{i\}) & \text{if } i \in S \\
                    \nu(S \cup \{i\}) - v_S & \text{otherwise}
                \end{cases}$
                \STATE $\hat\phi_i \leftarrow \frac{(M_i-1) \cdot \hat\phi_i + \Delta_i}{M_i}$
                \STATE $M_i \leftarrow M_i+1$
                \STATE $t \leftarrow t+1$
            \ENDFOR
            \STATE $M_{i,j} \leftarrow M_{i,j}+1$ for all $i,j \in P$
            \STATE $\Sigma_{i,j} \leftarrow \Sigma_{i,j} + (\Delta_i-\Delta_j)$ for all $i,j \in P$
            \STATE $\Gamma_{i,j} \leftarrow \Gamma_{i,j} + (\Delta_i-\Delta_j)^2$ for all $i,j \in P$
        \ENDWHILE
    \end{algorithmic}
    \textbf{Output}: $\hat{\mathcal{K}}$ containing $k$ players with highest estimate $\hat\phi_i$
\end{algorithm}
\begin{itemize}
    \item Initialize estimator $\hat\phi_i$ and individual counter of sampled marginal contributions $M_i$ for each player.
    \item Initialize for each player pair: the counter for joint appearances in rounds $M_{i,j}$, the sum of differences of marginal contributions $\Sigma_{i,j}$, and the sum of squared differences of marginal contributions $\Gamma_{i,j}$.
    \item Given $d_m := \Delta_i(S_m \setminus \{i\}) -\Delta_j(S_m \setminus \{j\})$ the unbiased variance estimator is 
    \begin{equation*}
        \begin{array}{l}
            \hat\sigma_{i,j}^2 := \frac{1}{M_{i,j}-1} \sum\limits_{m=1}^{M_{i,j}}(d_m-\bar{d})^2 = \frac{1}{M_{i,j}-1} \left(\Gamma_{i,j}-\frac{\Sigma_{i,j}^2}{M_{i,j}} \right) \, .
    	\end{array}
    \end{equation*}
    \item In each round, select with \textsc{\texttt{SelectPlayers}} players $P$ for whom to form an extended marginal contribution:
    \begin{itemize}
        \item First phase: select all players $M_\text{min}$ times: $P = \mathcal{N}$.
        \item Second phase: otherwise, partition the players into top-$k$ players $\hat{\mathcal{K}}$ and the rest $\hat{\mathcal{K}}' = \mathcal{N} \setminus \hat{\mathcal{K}}$ based on the estimates $\hat\phi_1,\ldots,\hat\phi_n$.
        \item Compute $\hat{p}_{i,j} \approx P(\phi_i < \phi_j)$ for all pairs $i \in \hat{\mathcal{K}}, j \in \hat{\mathcal{K}}'$.
        \item If all pairs are equally probable, select all players as it is not reasonable to be selective.
        \item Otherwise, sample a set of pairs $Q$ based on $\hat{p}_{i,j}$.
        \item Select all players as members of $P$ that are in at least one pair in $Q$.
    \end{itemize}
    \item Sample a coalition $S$ and cache its value.
    \item Form for all selected players in $P$ their extended marginal contribution $\Delta_i'(S)$ and update their estimator $\hat\phi_i$.
    \item Update the values $M_{i,j}$, $\Sigma_{i,j}$, and $\Gamma_{i,j}$ for all $i,j \in P$ required for computing the variance estimates $\hat{\sigma}_{i,j}^2$ and $\hat{p}_{i,j}$.
    \item In practice, we precompute and cache $\nu(\emptyset)$ and $\nu(\mathcal{N})$ in the beginning. We do that for \textbf{ALL} tested algorithms for a fair comparison.
    \item We modify Stratified SVARM to only precompute coalition values for sizes $0$ and $n$, instead of including sizes $1$ and $n-1$.
    Instead of integrating this optimization into all our algorithms, we remove it as it requires a budget of $2n$ which might be infeasible for games with large numbers of players.
\end{itemize}
\begin{algorithm}
    \caption{\textsc{\texttt{SelectPlayers}}}
    \label{alg:SelectPlayers}
    \begin{algorithmic}[1]
        \STATE $P \leftarrow \mathcal{N}$
        \IF{$M_{i,j} \geq M_\text{min}$ for all $i,j \in \mathcal{N}$}
            \STATE $\hat{\mathcal{K}} \leftarrow$ $k$ players of $\mathcal{N}$ with highest estimate $\hat\phi_i$, solve ties arbitrarily
            \STATE $\hat{\mathcal{K}}' \leftarrow \mathcal{N} \setminus \hat{\mathcal{K}}$
            \STATE $\hat\sigma^2_{i,j} \leftarrow \frac{1}{M_{i,j}-1} \left(\Gamma_{i,j} - \frac{\Sigma_{i,j}^2}{M_{i,j}}\right)$ for all $i \in \hat{\mathcal{K}}, j \in \hat{\mathcal{K}}'$
            \STATE $\hat{p}_{i,j} \leftarrow \Phi\left( \sqrt{M_{i,j}} \frac{-\Sigma_{i,j}}{\sqrt{\hat\sigma^2_{i,j}}}\right)$ for all $i \in \hat{\mathcal{K}}, j \in \hat{\mathcal{K}}'$ 
            \IF{$\min_{i,j} \hat{p}_{i,j} \ne \max_{i,j} \hat{p}_{i,j}$}
                \STATE $P, Q \leftarrow \emptyset$
                \FOR{$(i,j) \in \hat{\mathcal{K}} \times \hat{\mathcal{K}}'$}
                    \STATE Draw Bernoulli realization $B_{i,j}$ with $\mathbb{P}(B_{i,j}=1) = \frac{\hat{p}_{i,j}-\min_{i,j} \hat{p}_{i,j}}{\max_{i,j} \hat{p}_{i,j}-\min_{i,j} \hat{p}_{i,j}}$
                    \IF{$B_{i,j} = 1$}
                        \STATE $Q \leftarrow Q \cup \{(i,j)\}$
                        \STATE $P \leftarrow P \cup \{i,j\}$
                    \ENDIF
                \ENDFOR
            \ENDIF
        \ENDIF
    \end{algorithmic}
    \textbf{Output}: $\mathcal{P}$
\end{algorithm}

\newpage

\section{Further Empirical Results}
\label{app:more_empirical_results}

\begin{figure*}[h]
    \begin{center}
        \includegraphics[width=0.88\textwidth]{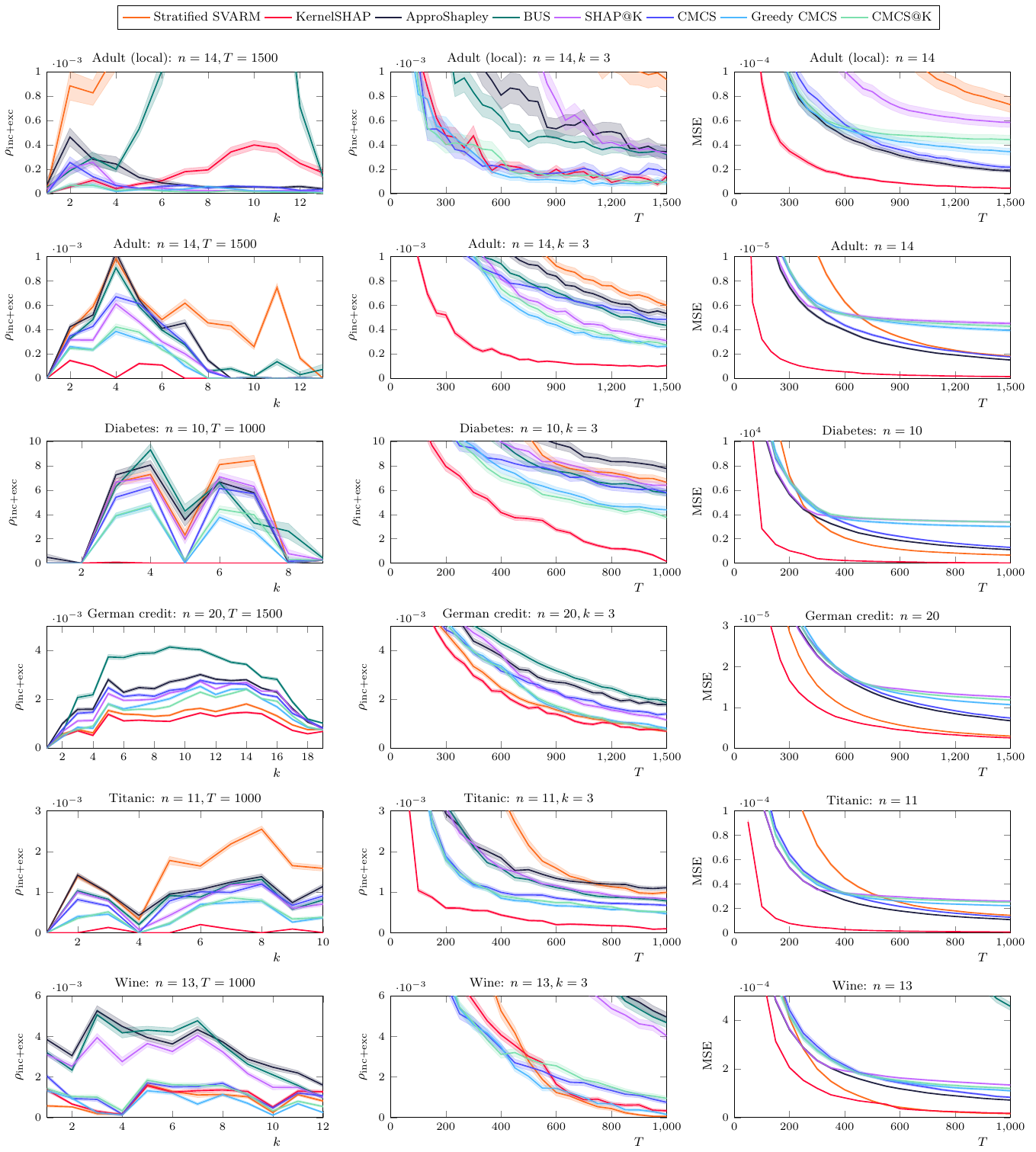}
        \vskip -0.15in
        \caption{
            Comparison of achieved error with baselines: inclusion-exclusion error for fixed budget with varying $k$ (left), inclusion-exclusion error for fixed $k$ with increasing budget (middle), and MSE depending on budget (right).
        }
        \label{fig:sota_epsilon_remaining}
    \end{center}
\end{figure*}

\begin{figure*}[h]
    \begin{center}
        \includegraphics[width=0.99\textwidth]{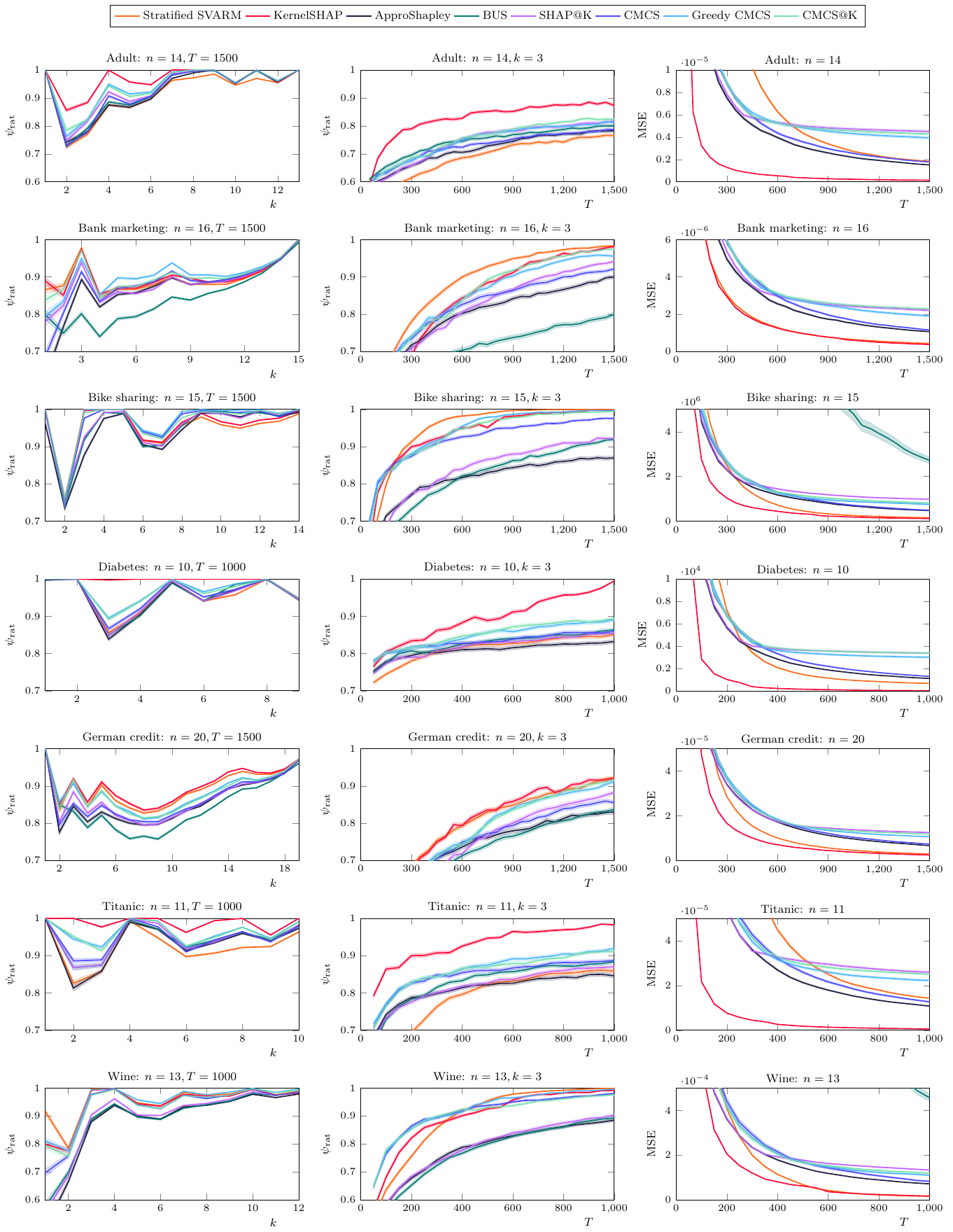}
        \vskip -0.15in
        \caption{
            Comparison of achieved ratio precision and MSE with baselines for global explanations: precision for fixed budget with varying $k$ (left), precision for fixed $k$ with increasing budget (middle), and MSE depending on budget (right).
        }
        \label{fig:sota_ratio_global}
    \end{center}
\end{figure*}

\begin{figure*}[h]
    \begin{center}
        \includegraphics[width=0.99\textwidth]{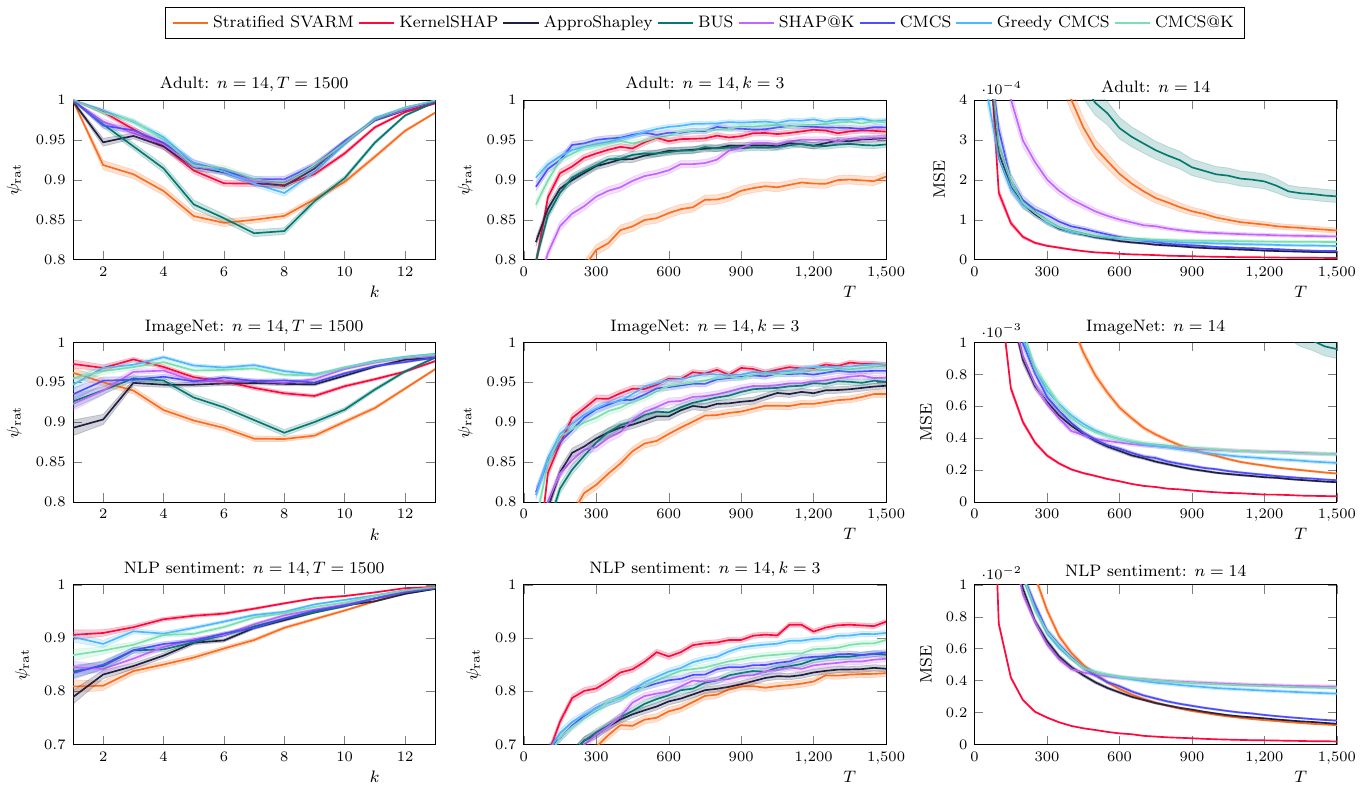}
        \vskip -0.15in
        \caption{
            Comparison of achieved ratio precision and MSE with baselines for local explanations: precision for fixed budget with varying $k$ (left), precision for fixed $k$ with increasing budget (middle), and MSE depending on budget (right).
        }
        \label{fig:sota_ratio_local}
    \end{center}
\end{figure*}

\end{document}